\documentclass[10pt,twocolumn,letterpaper]{article}

\usepackage{cvpr}              
\definecolor{cvprblue}{rgb}{0.21,0.49,0.74}
\usepackage[pagebackref,breaklinks,colorlinks,allcolors=cvprblue]{hyperref}
\usepackage{makecell}
\usepackage{bm}
\usepackage{multirow}
\usepackage{tabularx}
\usepackage{listings}

\def\paperID{15464} 
\def\confName{CVPR}
\def\confYear{2026}

\title{UCAgents: Unidirectional Convergence for Visual Evidence Anchored Multi-Agent Medical Decision-Making}

\newcommand*\samethanks[1][\value{footnote}]{\footnotemark[#1]}

\author{Qianhan Feng$\rm ^{1}$ , Zhongzhen Huang$\rm ^{2}$, Yakun Zhu$\rm ^{2,3}$ , Xiaofan Zhang$\rm ^{2,3}$\thanks{Corresponding Author}\space, Qi Dou$\rm ^{1}$\samethanks\\
$\rm ^{1}$The Chinese University of Hong Kong, Hong Kong SAR, China\\
$\rm ^{2}$Shanghai Jiao Tong University, Shanghai, China\\
$\rm ^{3}$Shanghai Innovation Institute, Shanghai, China\\
{\tt\small qianhan.feng@link.cuhk.edu.hk, qidou@cuhk.edu.hk}
}

\begin{document}
\maketitle
\begin{abstract}

Vision-Language Models (VLMs) show promise in medical diagnosis, yet suffer from reasoning detachment, where linguistically fluent explanations drift from verifiable image evidence, undermining clinical trust. Recent multi-agent frameworks simulate Multidisciplinary Team (MDT) debates to mitigate single-model bias, but open-ended discussions amplify textual noise and computational cost while failing to anchor reasoning to visual evidence, the cornerstone of medical decision-making. We propose \textbf{UCAgents}, a hierarchical multi-agent framework enforcing unidirectional convergence through structured evidence auditing. Inspired by clinical workflows, UCAgents forbids position changes and limits agent interactions to targeted evidence verification, suppressing rhetorical drift while amplifying visual signal extraction. In UCAgents, a one-round inquiry discussion is introduced to uncover potential risks of visual-textual misalignment. This design jointly constrains visual ambiguity and textual noise, a dual-noise bottleneck that we formalize via information theory. Extensive experiments on four medical VQA benchmarks show UCAgents achieves superior accuracy (71.3\% on PathVQA, +6.0\% over state-of-the-art) with 87.7\% lower token cost, the evaluation results further confirm that UCAgents strikes a balance between uncovering more visual evidence and avoiding confusing textual interference. These results demonstrate that UCAgents exhibits both diagnostic reliability and computational efficiency critical for real-world clinical deployment. Code is available at https://github.com/fqhank/UCAgents.

\end{abstract}    
\section{Introduction}

\begin{figure}[h]
\centering
\includegraphics[width=\columnwidth]{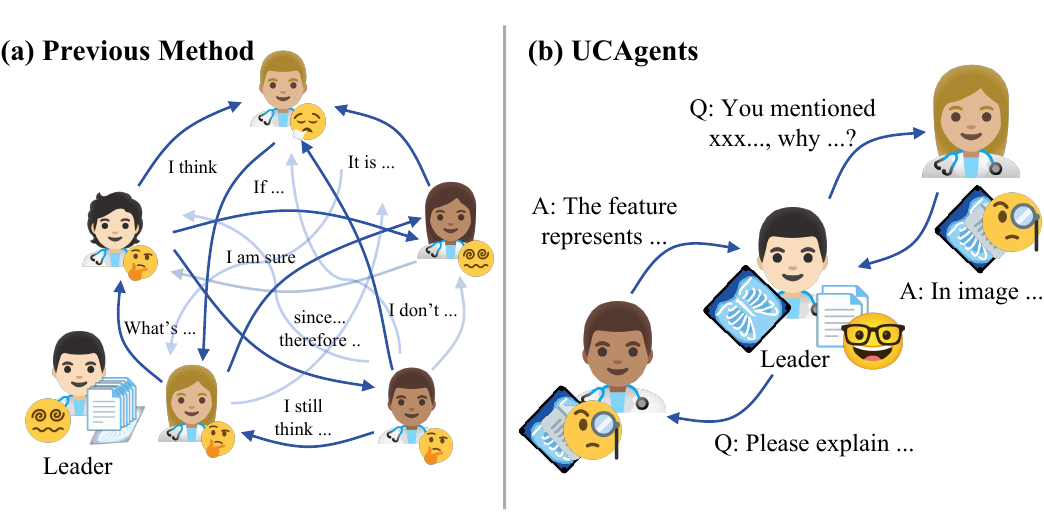} 
\caption{Unlike redundant discussion in previous multi-agent system, UCAgents uses one-round unidirectional inquiry to cut down textual noise and help focus on visual evidence.}
\label{fig:head}
\vspace{-0.2cm}
\end{figure}

Recent advances in Vision-Language Models (VLMs) \cite{gpt4,qwen2,chen2024internvlscalingvisionfoundation,llava,gemini,lu2024deepseek,yang2025qwen3technicalreport,zhang2024vision} have enabled unified reasoning across visual and textual modalities, achieving remarkable generalization in open-domain tasks. However, when applied to medical Visual Question Answering (VQA), a key benchmark for reliable AI-assisted diagnosis, their reliability sharply declines. The limitation lies in the fragile coupling between visual evidence and diagnostic reasoning. Even a subtle misinterpretation can reverse a clinical conclusion, exposing the need for reasoning that remains tightly anchored to verifiable image evidence. Medical reasoning fundamentally differs from general multimodal tasks: each inference step must be grounded in observable, clinically meaningful visual cues. Yet, current VLMs often produce linguistically fluent but visually unsupported explanations, namely reasoning detachment, where textual fluency conceals evidential drift and undermines clinical trust.

To address this, recent works have explored \textit{multi-agent collaboration} to simulate Multidisciplinary Team (MDT) discussions in clinical practice~\cite{kim2024mdagents,li2023camel}. While debate-based systems such as MDAgents~\cite{kim2024mdagents} encourage diverse reasoning, they suffer from an information–noise paradox: as discussion rounds increase, total information expands but textual noise $N_t$ inflates, leading to rhetorical overfitting and unstable convergence. Conversely, single-agent systems avoid excessive $N_t$ but amplify visual noise $N_v$ due to lack of cross-validation. These dual sources of uncertainty form the dual-noise bottleneck, where suppressing one inevitably worsens the other.

We propose \textbf{UCAgents} (Unidirectional Convergence Agents), a hierarchical multi-agent framework that replaces open debate with structured, entropy-reducing collaboration. UCAgents enforces a hierarchical flow of reasoning through three tiers:  
(1) independent agents perform controlled initial divergence to quantify uncertainties;  
(2) a supervisory reviewer audits visual-textual alignment to remove false consensus; and  
(3) a leader agent conducts evidence-anchored one-round risk inquiry by clarifying ambiguous findings and selecting the most visually consistent hypotheses. This process mirrors the clinical workflow of screening, verification, and adjudication, progressively filtering noise and consolidating diagnostic consensus around image-grounded evidence.

We evaluate UCAgents on several medical benchmarks, covering pathology, radiology, and mixed imaging modalities. Across both open-source and proprietary models, UCAgents consistently achieves superior visual-evidence anchored performance. Our main contributions lie in:
\begin{itemize}
    \item We identify and formalize the dual-noise bottleneck in medical VQA, revealing how visual ambiguity and textual drift jointly degrade diagnostic reliability.
    \item We propose a hierarchical, entropy-minimizing multi-agent framework that transforms open debates into unidirectional, evidence-anchored convergence.
    \item We demonstrate that UCAgents achieves consistent gains across four medical VQA benchmarks and various backbones, delivering clinically reliable reasoning with significantly improved accuracy and interpretability.
\end{itemize}

\section{Related Work}

\subsection{Medical Multimodal Reasoning and Visual Evidence Grounding}

Medical VQA has become a key benchmark for evaluating multimodal understanding in medical AI \cite{slake,vqa-rad,pathvqa,abacha2019vqa,zuo2025medxpertqa,wu2025bridge,Lin_2023}, requiring models to interpret fine-grained radiological features and integrate clinical queries into coherent reasoning \cite{pathvqa,vqa-rad,slake}. Datasets such as PathVQA \cite{pathvqa}, VQA-RAD \cite{vqa-rad}, and SLAKE \cite{slake} provide large-scale, clinically annotated image-question-answer pairs, lay the foundation for standardized evaluation across perception, interpretation, and diagnostic inference.

Recent advances in VLMs such as GPT-4 \cite{gpt4}, LLaVA \cite{llava}, Qwen-VL \cite{qwen2}, and Gemini \cite{gemini} have achieved impressive performance on open-domain multimodal reasoning. However, when applied to medical VQA, their visual encoders trained on natural images lack sensitivity to clinical features such as lesion texture, margin irregularity, or enhancement patterns \cite{lu2024visual}. They often produce linguistically fluent but visually unsupported reasoning, while domain-specific medical foundation models \cite{chen2024towards,ryu2025vision} capture fine-grained features but exhibit poor generalization across modalities or question types.

To improve reliability, researchers have explored three directions. Prompt-engineering strategies like Chain-of-Thought (CoT) and Self-Consistency (SC) \cite{DBLP:conf/nips/Wei0SBIXCLZ22,DBLP:conf/iclr/0002WSLCNCZ23} enhance logical structure but remain language-centric. Knowledge-augmented models \cite{singhal2023expertlevelmedicalquestionanswering} integrate medical ontologies but rarely verify consistency between reasoning and visual evidence. Despite progress, existing systems fail to jointly constrain visual ambiguity and textual redundancy, leaving diagnostic reasoning loosely anchored to image evidence.

\subsection{Multi-Agent Collaboration for Clinical Decision-Making}

Multi-agent systems inspired by Multidisciplinary Team collaboration \cite{kim2024mdagents,xia2025mmedagent,chen2025enhancing} have been proposed to enhance robustness through structured dialogue among specialized agents. Early frameworks such as CAMEL \cite{li2023camel} and AutoGen \cite{DBLP:journals/corr/abs-2308-08155} decompose tasks into fixed roles, improving interpretability but lacking agents explicitly responsible for verifying visual–textual alignment. Later works, MDAgents \cite{kim2024mdagents}, ReConcile \cite{DBLP:conf/acl/ChenSB24}, and Reflexion \cite{shinn2023reflexion}, introduced iterative debates or self-reflection to refine consensus. While these systems simulate human-like deliberation, open-ended exchanges often amplify textual noise and rhetorical drift, causing reasoning to deviate from visual evidence. Dynamic optimization frameworks \cite{dylan,DBLP:journals/corr/abs-2401-12954} further adapt team size or communication depth based on task complexity, yet focus mainly on interaction efficiency rather than maintaining evidence-centered reasoning.

In summary, current multi-agent methods oscillate between rigid role assignments that limit perspective diversity and unconstrained debates that increase linguistic entropy. Few explicitly model how multimodal evidence should be verified and filtered during collaboration. Our framework, UCAgents, addresses this gap through hierarchical, unidirectional convergence, enforcing structured information flow that stabilizes reasoning and anchors collaboration to verifiable visual evidence.

\section{Method: UCAgents}

\begin{figure*}[t]
\centering
\includegraphics[width=\textwidth]{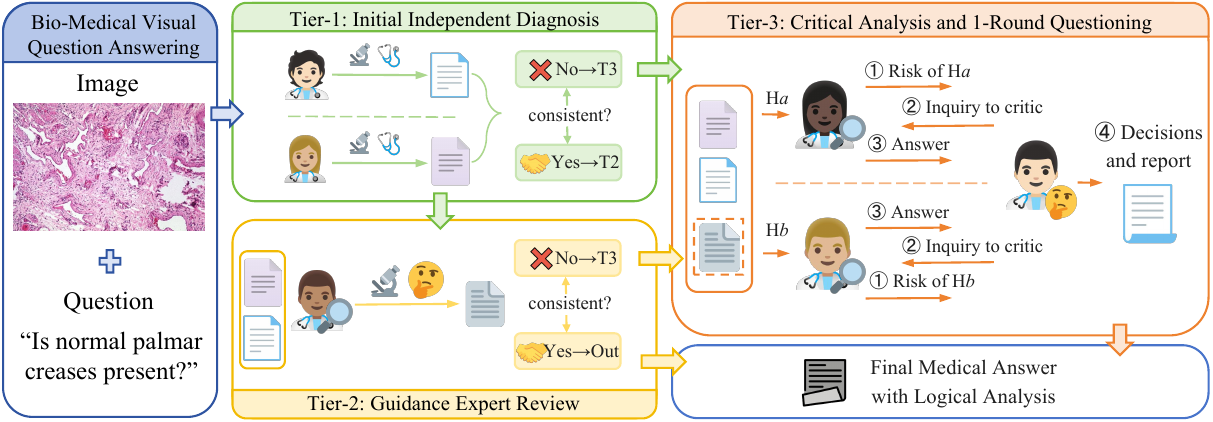} 
\caption{The overview of UCAgents. UCAgents system is composed of 3 dynamic Tiers: Initial Independent Diagnosis, Guidance Expert Review and Critical Analysis and Questioning. $H_{a}$, $H_{b}$: Divergent
Candidate Hypotheses from previous tiers.}
\label{fig:main}
\vspace{-0.3cm}
\end{figure*}

\subsection{Problem Formulation and Motivation}

Medical Visual Question Answering is a critical evidence-driven task where diagnostic reliability depends on how well reasoning anchors to medical images, the core evidence in multimodal inputs. We formalize this task through an information-theoretic lens to reveal performance bottlenecks in existing methods and motivate our structured multi-agent framework.

\textbf{Formalizing Medical VQA: The Primacy of Visual-Evidence Alignment}. Given a medical image $V$, clinical query $T$, the goal is to extract core visual evidence according to the textual constraint, and predict the correct answer $Y$ from the diagnostic hypothesis space $\mathcal{H}$. According to Fano's Inequality \cite{verdu1994generalizing}, the diagnostic error rate $P_e$ is bounded by:
\begin{equation}
\label{fano}
P_e \geq \frac{H(Y) - I(Y; \mathcal{I}) - 1}{\log |\mathcal{H}|},
\end{equation}
where $H(Y)$ is the prior entropy of the answer space, and $\mathcal{I} = (V, T, M)$ represents total observable information, visual evidence $V$, textual query $T$, and auxiliary information $M$ (e.g., agent interactions). The chain rule of mutual information yields:
\begin{equation}
I(Y; V, T) = I(Y; V) + I(Y; T|V),
\end{equation}
where $I(Y; V)$ captures core evidence association and $I(Y; T|V)$ captures task-guided visual reasoning. This leads to a critical insight: to minimize diagnostic errors, we must maximize $I(Y; \mathcal{I})$ by prioritizing $I(Y; V)$ while enhancing $I(Y; T|V)$, aligning with medical VQA's evidence-centric nature where diagnoses are ultimately grounded in images.

Unlike natural image tasks, Medical VQA is more difficult for VLMs because of the scarcity of high-quality medical data, stringent demands for accurate visual-textual alignment, and the high-stakes nature of medical decision-making. Single-agent VLMs for Medical VQA suppress $I(Y; \mathcal{I})$ through inadequate multimodal alignment:
\begin{itemize}
    \item \textbf{Unaudited visual-textual misalignment}: Small perceptual errors (e.g., misidentifying benign lesions) propagate unchallenged, reducing $I(Y; V)$ and breaking the link between textual arguments and visual features.
    \item \textbf{Single-point bias}: Lack of cross-validation leads to model biases (e.g., overdiagnosing common conditions), limiting $I(Y; T|V)$ as reasoning drifts from task-specific queries to biased priors.
\end{itemize}

Recent multi-agent frameworks inspired by MDT debates allow open, multi-round discussions. While designed to enhance $I(Y; \mathcal{I})$, they introduce detrimental entropy inflation that degrades cross-modal alignment:

\begin{itemize}
    \item \textbf{Unconstrained decision entropy}: Agents freely revise positions and explore open-ended justifications, expanding conditional entropy $H(Y|A_i)$. For medical VQA where answers are uniquely anchored to $V$, this diverts attention from visual-textual alignment to rhetorical persuasion.
    
    \item \textbf{Mutual information interference}: Lengthy debates introduce redundant textual noise, making $I(Y; V, M) \leq I(Y; V)$. Instead of enhancing alignment, auxiliary information $M$ creates distraction: $I(Y; M|V) > 0$ but often correlates with incorrect justifications.
    
    \item \textbf{Group consensus bias}: Open debates enable ``herd behavior'': agents with correct thinking may be swayed by majority incorrect reasoning, suppressing $I(Y; V)$.
    
    \item \textbf{Prohibitive computational cost}: Open debates require extensive tokens, but entropy inflation means cost doesn't translate to proportional performance gains.
\end{itemize}

\textbf{Principles of UCAgents}. Previous failures point to a clear principle: multi-agent collaboration must be structured to serve multimodal alignment and evidence-centricity, not open-ended debates. To this end, we propose a new multi-agent system for Medical VQA, UCAgents, which is demonstrated in Fig. \ref{fig:main}, achieves three complementary goals:

\begin{itemize}
    \item \textbf{Enhance visual-textual alignment}: Leverage multi-agent collaboration to audit and refine visual feature extraction directly boosting $I(Y; V)$ and $I(Y; T|V)$.
    \item \textbf{Suppress irrelevant textual noise}: Constrain interactions to avoid open debates, ensuring exchanges target validating visual evidence.
    \item \textbf{Eliminate group consensus bias}: Fix agent roles and enforce unidirectional convergence so reasoning relies solely on visual-textual evidence. 
\end{itemize}

\subsection{Overview of UCAgents Framework}

UCAgents is a dynamic 3-Tiers system as shown in Fig. \ref{fig:main}, it establishes a hierachical unidirectional process: \textbf{diversify} $\rightarrow$ \textbf{verify} $\rightarrow$ \textbf{converge}. Each Tier systematically reduces irrelevant entropy and reinforces visual grounding through fixed roles and structured interactions. In UCAgents, no one single agent is allowed to change their stances, and the opinion information provided to each one is strictly limited, thus to reduce the distraction from the image. We introduce each Tier and the special mechanism of UCAgents in detail in the following sections.

\subsection{Tier-1: Initial Independent Diagnosis}

Tier-1 objectively assesses the given medical case through structured collaboration. MDAgents \cite{kim2024mdagents} uses a single agent to evaluate the difficulty of the case, thus to determine whether the case is given to a single expert agent to process or to a MDT group to discuss. However, this highly subjective and biased process often leads to inappropriate case allocations. In UCAgents, we assess the case by conducting parallel independent diagnosis: two identical expert agents $\mathcal{A}_{1-1}$ and $\mathcal{A}_{1-2}$ are deployed to answer the question based on input image respectively. There is no communication between agents, thus to ensure reasoning anchors to original input $V$ and $T$, and divergence arises solely from independent visual interpretation.

It is important to encourage diversity between the two agents in their thinking. Classical operations include image data augmentations, but this would inevitably increase the difficulty of interpreting medical image which is already quite noisy. Some other method let different agents focus on different perspective like global versus local part. However, this operation explicitly ignores the textual needs, while most medical case need to combine both local and global information to give the correct diagnosis. To generate diversity without compromising image integrity, we introduce asymmetry via temperature modulation ($\tau = 0.7$). This slightly higher temperature encourages exploration of subtle visual variations without drifting into unconstrained textual reasoning. Since agents share identical setups, divergence amplifies inherent image ambiguity (e.g., ill-defined lesion boundaries) rather than introducing false differences. Each agent outputs an initial report that includes:  
\begin{itemize}
    \item $H_{1-i}$: diagnostic hypothesis.
    \item $R_{1-i}$: brief textual justification tied to visual features.
\end{itemize}

The two independent initial reports, $H_{1-1}$ and $H_{1-2}$, are compared: $D = \mathbb{I}[H_{1-1} \neq H_{1-2}]$. Disagreement $D = 1$ signals high divergence and difficulty requiring further in-depth discussions from experienced experts. However, consensus $D = 0$ cannot indicate reliable diagnosis, but may hide shared biases needing verification. We allocate the case into different reviewing routes based on $D$: 

\begin{equation}
\text{Route}(H_{1-1}, H_{1-2}) = \begin{cases} 
\text{Tier-3}, &D = 1,\\ 
\text{Tier-2}, &D = 0. 
\end{cases}
\end{equation}

Tier-1's entropy partitioning operator $\mathcal{P}_1$ formalizes the mutual information division:
\begin{equation}
I_1(Y; V,T) = I(Y; V,T | D=1) + I(Y; V,T | D=0),
\end{equation}
where $I(Y; V,T | D=1)$ and $I(Y; V,T | D=0)$ represent high-entropy and low-entropy cases respectively. The no-interaction design preserves $\mathcal{I} = (V, T)$ and avoids textual noise, while dual-agent reasoning enhances $I(Y; V)$ by double-mining visual evidence.

\subsection{Tier-2: Consensus Purification via Visual-Textual Alignment Verification}

Tier-2 tries to address the potential ``false consensus'', where Tier-1 agents agree due to shared biases rather than accurate visual-textual alignment. Its mission is to verify consensus authenticity, filter misaligned conclusions, and either finalize reliable diagnoses or escalate complex cases.


Tier-2 deploys a single Guidance Expert Agent $\mathcal{A}_2$, analogous to a supervising physician. $\mathcal{A}_2$ specializes in bidirectional validation - confirming whether consensus anchors to visual evidence and identifying potential flaws. It operates at moderate temperature ($\tau = 0.5$) for stability and rigorous evidence checking.

The agent's input materials leverages Tier-1's outputs: original medical image $V$, clinical query $T$, and consensus report from Tier-1 (including $H_1=H_{1-1}=H_{1-2}$ and justifications $R_{1-1}, R_{1-2}$). Its structured workflow is formulated as:

\begin{enumerate}
    \item \textbf{Comprehensive Visual Scan}: Identify core diagnostic features in $V$ (e.g., lesion shape, density), but without providing preconceived judgments or positions. This step solely helps Guidance Expert have an initial knowledge of the case.  
    \item \textbf{Evidence verification}: Parse $R_{1-1}$ and $R_{1-2}$ to verify each claim aligns with visual features (eliminating hallucinations) and no critical features are omitted.
    \item \textbf{Logic verification}: Parse $R_{1-1}$ and $R_{1-2}$ to check whether the visual evidence is correctly interpreted, and whether the logical reasoning in the reports is rigorous and correct.
    \item \textbf{Independent hypothesis generation}: Produce a new diagnosis $H_2$ and report $R_2$ based on extracted verification rather than independent reasoning.
\end{enumerate}

The consensus purification operator $\mathcal{P}_2$ in Tier-2 is:
\begin{equation}
I_2(Y; V,T | D=0) = I(Y; V,T | H_2=H_1),
\end{equation}
Guidance Expert gains access to more ideas and information while conducting reviews, but this information is neither disordered nor noisy. Instead, it is confined to different lines of reasoning that lead to the same conclusion. When the expert’s conclusion $H_2$ aligns with that of Tier-1 $H_1$, it indicates the conclusion has high reliability. However, if they identify flaws after reviewing Tier-1’s conclusion and propose an alternative solution, this signifies the case is quite high in complexity and requires further research:
\begin{equation}
\text{Route}(H_1, H_2) = \begin{cases} 
\text{Terminates}, & H_2 = H_1,\\ 
\text{Tier-3},  & H_2 \neq H_1. 
\end{cases}
\end{equation}

Consensus is finalized only if hypotheses agree and alignment is verified, ensuring conclusions are correct for the right reasons. Disagreement or detected misalignment (e.g., unsubstantiated claims, omitted features) triggers escalation. This enhances mutual information: $I_2(Y; V, T|D=0) > I_1(Y; V, T|D=0)$ by filtering false consensus derived from shared biases.

\subsection{Tier-3: Unidirectional Risk Auditing with Targeted Inquiry}

Building upon the structured uncertainty estimation (Tier-1) and bias correction (Tier-2), Tier-3 conducts final adversarial risk auditing to ensure convergence toward evidence-grounded diagnosis. Facing this challenge, Tier-3 deploys specialized agents with fixed, non-overlapping roles:

\begin{itemize}
    \item \textbf{2 Critical Analyst Agents} ($\mathcal{C}_1, \mathcal{C}_2$): Examine the potential risk of existing hypotheses. 
    \item \textbf{1 Leader Agent} ($\mathcal{A}_L$): Acts as the MDT chairperson who facilitates discussion and makes the final arbitration. 
\end{itemize}
By organizing a discussion with these agents, Tier-3 operates via a four-step processto establish the final diagnosis.

\textbf{Step 1: Unidirectional Risk Mining.} Both $\mathcal{C}_1$ and $\mathcal{C}_2$ are exclusively assigned to challenge 1 specific hypothesis. Unlike Tier-2's Guidance Expert who evaluates whether reasoning is sound, Critical Analysts adopt a \textbf{``devil's advocate'' stance}: actively mining potential flaws and risks that could invalidate the hypothesis with moderate 0.5 temperature.
Each analyst outputs a structured risk report:
\begin{equation}
\mathcal{C}(H_i, R_i, V, T) \rightarrow R^{risk}_{i},
\end{equation}
where $ R^{risk}_{i}$ contains identified flaws in analysis.

This mechanism has 2 advantages: (1) Opinions on each hypothesis are consolidated into one report respectively, avoiding herd behavior caused by discrepancies in the number of agents holding different views, (2) Targeted critical risk reports can offset the persuasive rhetoric of supporting opinions to a certain extent.

\textbf{Step 2\&3: Leader-Directed Inquiry and Expert Response.} After receiving both risk reports, the Leader Agent $\mathcal{A}_L$ reviews them to identify areas requiring deeper investigation. Unlike open debates where agents freely exchange arguments, the Leader issues a single targeted question $Q_i$ to each Critical Analyst to probe specific ambiguities or underdeveloped claims in their risk assessment. This inquiry operator is formalized as:
\begin{equation}
\hat{R}^{risk}_{i} = \mathcal{A}_L(R^{risk}_{i}, Q_i, V, T),
\end{equation}
where $\hat{R}^{risk}_{i}$ is the supplemented response. This step mirrors clinical practice where the MDT chairperson asks pointed questions to clarify conflicting interpretations without allowing open-ended debate. The inquiry $Q_i$ acts like applying attention to the logic and encourage reinforcement. If the original $H_{i}$ is wrong, $\hat{R}^{risk}_{i}$ would be more robust. In the contrast, if $H_{i}$ is correct, then $\hat{R}^{risk}_{i}$ would reveal more flaws in $R^{risk}_{i}$. Uniform 0.1 temperature is used.

\textbf{Step 4: Final Arbitration.} Given the response from both $\mathcal{C}_1$ and $\mathcal{C}_2$, Leader agent $\mathcal{A}_L$ is required to evaluate the overall risks by comparing the reports and aggregating all the information, and make the final diagnosis:
\begin{equation}
Y^* = \mathcal{A}_L(\hat{R}^{risk}_{a}, \hat{R}^{risk}_{b}, R, V, T),
\end{equation}
and the Leader is also allowed to make a diagnosis out of ${H_{a},H_{b}}$, if enough evidence suggests that both are risky and identified evidences lead to a new solution. 

The key innovation of Tier-3 lies in its unidirectional risk auditing mechanism: instead of having agents debate which diagnosis is correct, which may lead to rhetorical persuasion and entropy inflation, we task agents with actively searching for reasons why one specific hypothesis might be \textit{wrong}, and one-round inquiry-response communication helps the unidirectional convergence of the critical logics. This adversarial approach forces deep scrutiny of visual evidence while avoiding the pitfalls of open debate.

The unidirectional auditing maximizes mutual information $I(Y; V, T)$ while suppressing irrelevant entropy from textual debates $M$. The adversarial risk reports force agents to extract counter-evidence from $V$ that might invalidate each hypothesis, effectively amplifying signal from visual features that distinguish correct from incorrect diagnoses. By constraining communication to a single targeted inquiry per agent, we prevent the entropy inflation $H(Y|M)$ that plagues open debate systems, where lengthy arguments introduce noise orthogonal to visual evidence. Overall, The framework maximizes $I(Y; \mathcal{I})$ by prioritizing $I(Y; V)$ and $I(Y; T|V)$ while suppressing textual noise from $M$. 




\section{Experiments}

\subsection{Experimental Setup}

We conduct experiments on multiple medical VQA datasets and across various vision-language backbones. All experiments on open-source models are implemented under the Ollama framework \cite{ollama-search-2025} on a single NVIDIA H800 GPU, with results averaged over three independent trials. For fairness, we adopt a unified temperature configuration across all datasets and models. All agents are guided by standardized prompt templates, which can be found in appendix A to ensure consistent behavior across settings, without exemplar demonstrations. This guarantees reproducibility while minimizing prompt engineering bias, enabling a clean comparison between different multi-agent paradigms. All datasets used are open-source, and the research has obtained the necessary licenses and certifications.

\subsection{Results based on GPT-4}

We first evaluate UCAgents using \textbf{GPT-4} \cite{gpt4} as the base model. GPT-4 represents one of the advanced vision–language models, capable of joint image and text understanding across diverse tasks. We follow MDAgents to conduct experiments on two representative medical VQA benchmarks: \textbf{PathVQA}~\cite{pathvqa} and \textbf{MIMIC-CXR-VQA}~\cite{mimic}. PathVQA focuses on pathology slides involving fine-grained questions on tissue structure, cell morphology, and diagnostic patterns; MIMIC-CXR-VQA contains clinical chest X-rays paired with textual questions covering thoracic diseases such as pneumonia, effusion, and cardiomegaly. These two datasets collectively span both microscopic and radiographic diagnostic scenarios.

As shown in Table~\ref{table:exp1}, UCAgents achieves 71.3\% on PathVQA and 60.3\% on MIMIC-CXR-VQA, consistently outperforming all baselines. Compared to the best single-agent baseline, UCAgents yields a 9.9\% improvement. Among multi-agent methods, UCAgents surpasses the MDT-based MDAgents by 6.0\%, confirming the effectiveness of unidirectional convergence mechanism in maintaining visual-textual alignment and diagnostic stability. These results demonstrate that even with a strong base model like GPT-4, structured and directed collaboration remains critical for reliable medical reasoning, validating UCAgents as a robust enhancement over existing multi-agent paradigms.

\begin{table}[t]
\centering
\caption{Accuracy Results (\%) based on GPT-4 model.}
\begin{tabular}{@{}cccc@{}}
\toprule
Category & Method & Path-VQA & MIMIC-CXR \\
\midrule
\multirow{6}{*}{\makecell{Single\\Agent}} & Zero-Shot & 57.9\textsubscript{\textcolor{gray}{\textpm1.6}} & 40.0\textsubscript{\textcolor{gray}{\textpm5.3}}\\
& Few-Shot & 57.5\textsubscript{\textcolor{gray}{\textpm4.5}} & 35.3\textsubscript{\textcolor{gray}{\textpm5.0}}\\
& +CoT\cite{DBLP:conf/nips/Wei0SBIXCLZ22} & 58.6\textsubscript{\textcolor{gray}{\textpm3.1}} & 36.2\textsubscript{\textcolor{gray}{\textpm5.2}}\\
& +CoT-SC\cite{DBLP:conf/iclr/0002WSLCNCZ23} & 61.2\textsubscript{\textcolor{gray}{\textpm2.1}} & 51.7\textsubscript{\textcolor{gray}{\textpm4.0}}\\
& ER\cite{DBLP:journals/corr/abs-2212-13138} & 61.4\textsubscript{\textcolor{gray}{\textpm4.1}} & 50.0\textsubscript{\textcolor{gray}{\textpm0.0}}\\
& MedPrompt\cite{DBLP:journals/corr/abs-2311-16452} & 59.2\textsubscript{\textcolor{gray}{\textpm5.7}} & 53.4\textsubscript{\textcolor{gray}{\textpm4.3}}\\
\midrule
\multirow{7}{*}{\makecell{Multi\\Agents}} & Reconcile\cite{DBLP:conf/acl/ChenSB24} & 57.5\textsubscript{\textcolor{gray}{\textpm3.3}} & 33.3\textsubscript{\textcolor{gray}{\textpm3.4}}\\
& AutoGen\cite{autogen} & 43.0\textsubscript{\textcolor{gray}{\textpm8.9}} & 43.3\textsubscript{\textcolor{gray}{\textpm8.2}}\\
& DyLAN\cite{dylan} & 41.3\textsubscript{\textcolor{gray}{\textpm1.2}} & 38.7\textsubscript{\textcolor{gray}{\textpm1.2}}\\
& MedAgents\cite{tang2023medagents} & 45.4\textsubscript{\textcolor{gray}{\textpm8.1}} & 43.3\textsubscript{\textcolor{gray}{\textpm7.0}}\\
& Meta-Prompt\cite{DBLP:journals/corr/abs-2401-12954} & 55.3\textsubscript{\textcolor{gray}{\textpm2.3}} & 42.0\textsubscript{\textcolor{gray}{\textpm4.0}}\\
& MDAgents\cite{kim2024mdagents} & 65.3\textsubscript{\textcolor{gray}{\textpm3.9}} & 55.9\textsubscript{\textcolor{gray}{\textpm9.1}}\\
& \textbf{UCAgents} & \textbf{71.3\textsubscript{\textcolor{gray}{\textpm1.3}}} & \textbf{60.3\textsubscript{\textcolor{gray}{\textpm0.5}}}\\

\bottomrule
\end{tabular}
\label{table:exp1}
\vspace{-0.2cm}
\end{table}

\begin{table*}[t]
    \centering
    \caption{Performance on open-source models, with accuracy reported in percentage (\%). For VQA-RAD and SLAKE-VQA, the performances are reported in the format of \textit{full-set}/\textit{closed-set}. The consumed tokens are reported in the format of Input/Output token numbers. 1: Qwen2.5VL-3B based MDAgents is not reported since the system classifies all cases into the `\textit{Basic}' category. 2: The running time of Qwen2.5VL-72B based MDAgents on Path-VQA and MIMIC-CXR-VQA largely surpasses affordable range.}
    \resizebox{0.98\textwidth}{!}{
    \begin{tabular}{c|cccc|c|cc}
        \toprule[1pt]
Method & Path-VQA & MIMIC-CXR-VQA & VQA-RAD & SLAKE-VQA & AVG.& API Calls & Tokens(K) \\
\midrule
\textbf{Qwen2.5VL-3B} & 37.00\textsubscript{\textcolor{gray}{\textpm16.26}} & 44.05\textsubscript{\textcolor{gray}{\textpm1.24}} & 40.16\textsubscript{\textcolor{gray}{\textpm7.98}}/43.03\textsubscript{\textcolor{gray}{\textpm8.62}} & 52.05\textsubscript{\textcolor{gray}{\textpm0.30}}/55.78\textsubscript{\textcolor{gray}{\textpm3.59}} & 45.35 & 1.00 & 1.02/0.06 \\
w/MDAgents$^{1}$ & - & - & - & - & - & - & - \\
w/UCAgents & \textbf{61.46}\textsubscript{\textcolor{gray}{\textpm1.67}} & \textbf{53.09}\textsubscript{\textcolor{gray}{\textpm2.80}} & \textbf{52.64}\textsubscript{\textcolor{gray}{\textpm1.46}}/\textbf{54.09}\textsubscript{\textcolor{gray}{\textpm1.82}} & \textbf{58.64}\textsubscript{\textcolor{gray}{\textpm2.25}}/\textbf{60.58}\textsubscript{\textcolor{gray}{\textpm1.00}} & \textbf{56.75} & 5.38 & 6.49/0.40 \\
\midrule
\textbf{Qwen2.5VL-7B} & 55.15\textsubscript{\textcolor{gray}{\textpm10.21}} & 52.61\textsubscript{\textcolor{gray}{\textpm2.28}} & 53.59\textsubscript{\textcolor{gray}{\textpm0.11}}/57.37\textsubscript{\textcolor{gray}{\textpm1.13}} & 59.66\textsubscript{\textcolor{gray}{\textpm0.78}}/60.71\textsubscript{\textcolor{gray}{\textpm0.21}} & 56.15 & 1.00 & 0.99/0.06 \\
w/MDAgents & 58.13\textsubscript{\textcolor{gray}{\textpm3.42}} & 47.23\textsubscript{\textcolor{gray}{\textpm2.63}} & 53.44\textsubscript{\textcolor{gray}{\textpm1.60}}/59.05\textsubscript{\textcolor{gray}{\textpm2.23}} & 57.60\textsubscript{\textcolor{gray}{\textpm8.60}}/\textbf{63.02}\textsubscript{\textcolor{gray}{\textpm1.79}} & 56.41 & 10.15 & 21.02/1.74 \\
w/UCAgents & \textbf{61.90}\textsubscript{\textcolor{gray}{\textpm0.62}} & \textbf{54.45}\textsubscript{\textcolor{gray}{\textpm2.55}} &  \textbf{56.12}\textsubscript{\textcolor{gray}{\textpm0.00}}/\textbf{61.35}\textsubscript{\textcolor{gray}{\textpm0.00}} & \textbf{61.45}\textsubscript{\textcolor{gray}{\textpm2.93}}/62.48\textsubscript{\textcolor{gray}{\textpm0.11}} & \textbf{59.63} & 4.49 & 5.97/0.45 \\
\midrule
\textbf{Qwen2.5VL-32B} & 63.12\textsubscript{\textcolor{gray}{\textpm0.08}} & 53.30\textsubscript{\textcolor{gray}{\textpm0.62}} & 62.37\textsubscript{\textcolor{gray}{\textpm1.91}}/65.12\textsubscript{\textcolor{gray}{\textpm0.31}} & 71.53\textsubscript{\textcolor{gray}{\textpm1.88}}/69.72\textsubscript{\textcolor{gray}{\textpm0.59}} & 64.19 & 1.00 & 0.95/0.08 \\
w/MDAgents & 61.80\textsubscript{\textcolor{gray}{\textpm3.13}} & 51.68\textsubscript{\textcolor{gray}{\textpm0.28}} & 60.59\textsubscript{\textcolor{gray}{\textpm2.06}}/65.46\textsubscript{\textcolor{gray}{\textpm1.90}} & 73.13\textsubscript{\textcolor{gray}{\textpm1.48}}/69.24\textsubscript{\textcolor{gray}{\textpm2.69}} & 63.65 & 8.43 & 23.76/2.23 \\
w/UCAgents & \textbf{63.96}\textsubscript{\textcolor{gray}{\textpm0.28}} &\textbf{54.14}\textsubscript{\textcolor{gray}{\textpm0.40}} & \textbf{66.34}\textsubscript{\textcolor{gray}{\textpm1.73}}/\textbf{67.60}\textsubscript{\textcolor{gray}{\textpm0.57}} & \textbf{73.63}\textsubscript{\textcolor{gray}{\textpm1.18}}/\textbf{73.60}\textsubscript{\textcolor{gray}{\textpm1.71}} & \textbf{66.55} & 4.10 & 5.52/0.37 \\
\midrule
\textbf{Qwen2.5VL-72B} & 63.43\textsubscript{\textcolor{gray}{\textpm3.60}} & 56.36\textsubscript{\textcolor{gray}{\textpm2.44}} &65.49\textsubscript{\textcolor{gray}{\textpm1.78}}/74.91\textsubscript{\textcolor{gray}{\textpm1.69}} &73.22\textsubscript{\textcolor{gray}{\textpm0.16}}/74.61\textsubscript{\textcolor{gray}{\textpm1.25}} & 68.00 & 1.00 & 0.97/0.06 \\
w/MDAgents$^{2}$ & - & - & 64.28\textsubscript{\textcolor{gray}{\textpm2.28}}/70.31\textsubscript{\textcolor{gray}{\textpm2.68}} & \textbf{75.22}\textsubscript{\textcolor{gray}{\textpm1.34}}/72.82\textsubscript{\textcolor{gray}{\textpm2.38}} & - & 6.87 & 14.75/1.61 \\
w/UCAgents & \textbf{67.61}\textsubscript{\textcolor{gray}{\textpm0.57}} & \textbf{62.55}\textsubscript{\textcolor{gray}{\textpm1.24}} & \textbf{68.42}\textsubscript{\textcolor{gray}{\textpm0.16}}/\textbf{75.67}\textsubscript{\textcolor{gray}{\textpm1.65}} & 74.69\textsubscript{\textcolor{gray}{\textpm0.89}}/\textbf{80.04}\textsubscript{\textcolor{gray}{\textpm1.25}} & \textbf{71.50} & 3.71 & 4.78/0.32 \\
\midrule
\textbf{LLaVA-7B} & 47.78\textsubscript{\textcolor{gray}{\textpm1.23}} &44.05\textsubscript{\textcolor{gray}{\textpm1.87}} & 44.93\textsubscript{\textcolor{gray}{\textpm1.67}}/47.30\textsubscript{\textcolor{gray}{\textpm4.67}} & 47.08\textsubscript{\textcolor{gray}{\textpm2.47}}/46.69\textsubscript{\textcolor{gray}{\textpm0.69}} & 46.31 & 1.00 & 0.84/0.07 \\
w/MDAgents & 50.49\textsubscript{\textcolor{gray}{\textpm5.25}} & 44.23\textsubscript{\textcolor{gray}{\textpm3.43}} & 47.72\textsubscript{\textcolor{gray}{\textpm5.30}}/47.87\textsubscript{\textcolor{gray}{\textpm1.60}} & 43.84\textsubscript{\textcolor{gray}{\textpm1.89}}/46.05\textsubscript{\textcolor{gray}{\textpm1.20}} & 46.70 & 13.39 & 33.57/2.86 \\
w/UCAgents & \textbf{54.56}\textsubscript{\textcolor{gray}{\textpm1.22}} & \textbf{54.63}\textsubscript{\textcolor{gray}{\textpm1.53}} & \textbf{61.03}\textsubscript{\textcolor{gray}{\textpm3.14}}/\textbf{54.16}\textsubscript{\textcolor{gray}{\textpm0.89}}
& \textbf{60.03}\textsubscript{\textcolor{gray}{\textpm1.58}}/\textbf{50.04}\textsubscript{\textcolor{gray}{\textpm0.95}} & \textbf{55.74} & 4.48 & 5.78/0.45 \\
\bottomrule
\end{tabular}}
\label{table:exp2}
\vspace{-0.2cm}
\end{table*}

\subsection{Results based on Open-Source Models}

We further extend experiments to open-source VLMs including \textbf{Qwen2.5VL} \cite{qwen2} series and \textbf{LLaVA} \cite{llava}. In addition to the PathVQA and MIMIC-CXR-VQA, we include two complementary benchmarks, \textbf{VQA-RAD}~\cite{vqa-rad} and \textbf{SLAKE-VQA}~\cite{slake}. VQA-RAD integrates radiology images across CT, MRI, and X-ray modalities, while SLAKE-VQA focuses on anatomical reasoning and multilingual comprehension. Both datasets include a mixture of \textit{closed-form} and \textit{open-ended} questions. For the open-ended subset, we follow a standardized setup: GPT-4 is used to generate 3–4 plausible answer options including the correct one, ensuring balanced difficulty and fair comparison. We thus report both \textit{full-set}/\textit{closed-set} scores per dataset.

As shown in Table~\ref{table:exp2}, UCAgents consistently outperforms both single-agent and SOTA multi-agent method MDAgents across all settings. Notably, on Qwen2.5VL-3B, UCAgents achieves an average gain of 11.4\%, while on the larger 72B model, it still provides a solid 3.5\% improvement. On SLAKE-VQA, UCAgents reaches up to 80.0\%. Meanwhile, on VQA-RAD, UCAgents delivers +8.5\% and +10.6\% gains on the total and closed sets respectively, further validating its consistency across question types. What's more, with Qwen2.5VL-3B, MDAgents classifies almost all cases into `\textit{Basic}' category, thus performing equally as single agent.  

These results validate the proposed unidirectional convergence strategy’s generalization across architectures and modalities. Critically, UCAgents turns lightweight open-source VLMs into clinically reliable assistants while boosting lightweight models to match larger models’ diagnostic performance and enhancing powerful models’ stability by reducing dangerous misdiagnoses.

\subsection{Ablation Study}

We evaluate several key components proposed in UCAgents, and the results are reported in Table \ref{table:ablation}, which is conducted based on LLaVA-7B model on VQA-RAD dataset. The supervisor agent in Tier-2 provides further review on potential false consensus from Tier-1, and provides a 3.54\% accuracy gain to the final performance. In Tier-3, which is the core mechanism of UCAgents, there are 3 important methods ensuring the success of discussion: (1) one-round inquiry, (2) 2 independent Critical Analyst Agents and (3) unidirectional critics instead of support. In Table \ref{table:ablation}, removing the leader’s questioning step reduces overall performance by 15.60\%, and an extreme drop of 27.23\% on the accuracy of Tier-3 expert review can also be witnessed. Also, using one single agent (instead of two independent agents) to criticize both hypotheses reduces accuracy by 4.58\%, since the bias is inevitable. If the agents are instructed to provide supportive reports on hypotheses instead of finding risks, the discussion quality also suffers, which demonstrates that unidirectional critic is important for mitigating the eloquence effect. 
To further validate UCAgents’ effectiveness, we analyze the performance of different diagnostic routes. According to the GPT-4 based experiment on Path-VQA in Table \ref{table:routes}, the validated consensus from supervisor in Tier-2 achieves an accuracy of 73.47\%. For complex cases where disagreements arose at the Tier-1 level, the accuracy rate improved by 11.64\% after consultation with Tier-3 experts, compared to using a single agent to make a diagnosis directly (49.48\%). Identifying consensus risks in Tier-2 reviews often indicates an extremely high level of complexity in the cases, as evidenced by the poor performance of using a single agent to make decisions on these cases (43.03\%). However, through the efficient expert consultation of Tier-3, these cases can be handled with a high accuracy of 62.42\%.  

\begin{table}[t]
\centering
\caption{Ablation studies on VQA-RAD with LLaVA-7B model.}
\begin{tabular}{@{}ccc@{}}
\toprule
Tier & Method & Performance\\
\midrule
- & UCAgents & \textbf{61.03\%} \\
Tier-2 & w/o Supervisor Review & 57.49\%\\
Tier-3 & w/o One-Round Inquiry & 45.43\%\\
Tier-3 & w/o Independent Critics & 56.45\%\\
Tier-3 & w/o Critics \& w/ Support & 53.10\%\\
\midrule
- & UCAgents:Tier-3 & \textbf{63.21\%} \\
Tier-3 & w/o One-Round Inquiry & 35.98\%\\
Tier-3 & w/o Independent Critics & 59.07\%\\
Tier-3 & w/o Critics \& w/ Support & 49.76\%\\
\bottomrule
\end{tabular}
\label{table:ablation}
\vspace{-0.2cm}
\end{table}

\begin{figure*}[t]
\centering
\includegraphics[width=\textwidth]{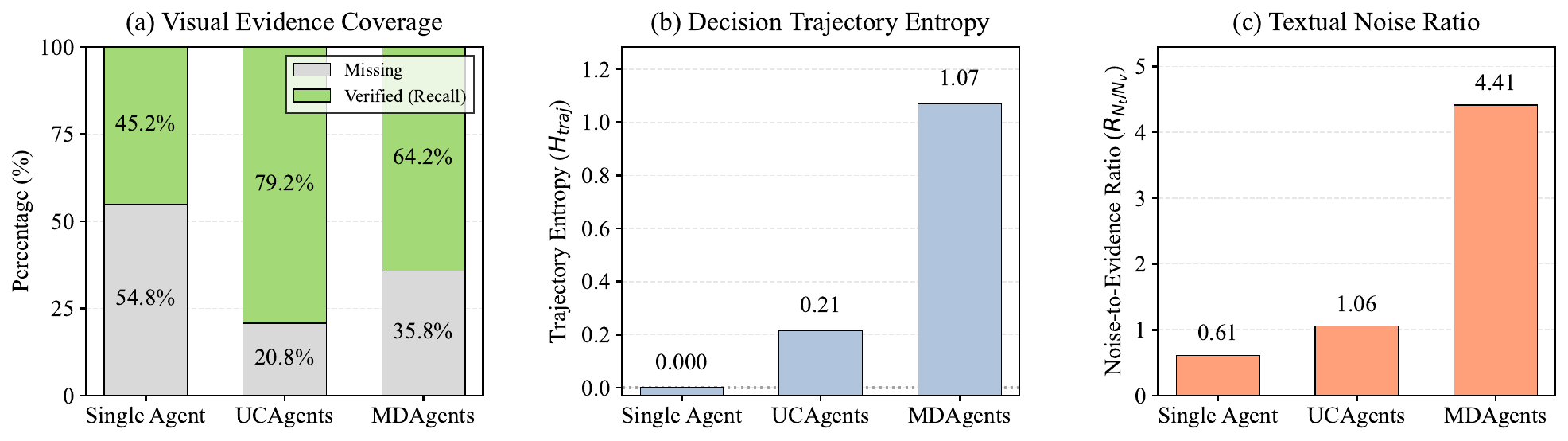} 
\caption{Visual-evidence anchored diagnosis quality analysis. (a) Visual Evidence Coverage. UCAgents recalls more verified visual evidence than MDAgents from the image. (b) Decision Trajectory Entropy. Unidirectional Covergence mechanism reduces agents' confusion caused by noisy decision space compared to MDAgents. (c) Textual Noise Ratio. UCAgents achieves a balance between evidence sentences and distractive sentences. Outer assistant processes 3 records together at one time for fairness.}
\label{fig:quality}
\vspace{-0.3cm}
\end{figure*}

\subsection{Visual-Anchored Diagnosis Quality Analysis}

To validate UCAgents's core design principles: visual-evidence anchored reasoning, unidirectional convergence and entropy control of the system, we conduct a three-dimensional quality analysis with LLaVA model on VQA-RAD dataset, comparing single agent, UCAgents and MDAgents. We use Gemini-2.5-pro \cite{gemini} as outer assistant. 

\textbf{Visual Evidence Coverage}. UCAgents enhances visual-textual alignment through hierarchical review and verification. We compute the average identified and missing key visual evidences across all samples. As shown in Fig. \ref{fig:quality}(a), UCAgents successfully recalls 79.2\% important visual evidences, while MDAgents misses 35.8\%. 

\textbf{Decision Trajectory Entropy}. We quantify decision stability by measuring the entropy of diagnostic hypotheses proposed during reasoning: 
\begin{equation}
\vspace{-0.2cm}
    H_{traj} = -\sum_{i=1}^{K} p(h_i) \log_2 p(h_i),
\end{equation}
where $K$ is the number of proposed hypotheses and $p(h_i)$ their frequency. Higher entropy indicates unstable, exploratory decision-making. According to Fig. \ref{fig:quality}(b) MDAgents shows high entropy
and 38.6\% producing more than 3 competing hypotheses. Contrastively, UCAgents exhibits controlled low entropy of 0.214. This near-zero entropy validates our ``Unidirectional Convergence" design. Conversely, MDAgents's entropy explosion confirms the failure mode: unconstrained debates expand $H(Y|M)$ as agents shift allegiances based on eloquence rather than evidence. 

\textbf{Textual Noise-to-Signal Ratio}. We compute the noise/visual-evidence sentences ratio $R_{N_{t}/N_{v}}$ by parsing the diagnostic records. Noise includes rhetorical statements, procedural comments, and persuasive arguments lacking evidence. This ratio quantifies ``textual entropy inflation": systems with high $R_{N_{t}/N_{v}}$ bury diagnostic signals under irrelevant verbiage. UCAgents exhibits low noise ratio of 1.06, which is at the same level of single agents, and MDAgents gives catastrophic noise of 4.41.

Fig.\ref{fig:quality} reveals that even though MDAgents excavate more visual evidences than single agent, its chaotic communication mechanism and highly disordered system have offset this advantage. UCAgents achieves a remarkable balance and keeps the attention on core visual evidence.

\begin{table}[t]
\centering
\caption{Detailed accuracy of different diagnosis routes on Path-VQA, based on GPT-4. T stands for Tier.}
\begin{tabular}{@{}ccc@{}}
\toprule
Route & Accuracy & w/Single-Agent\\
\midrule
T1$\rightarrow$T2 & 73.47\% & 61.43\%($\downarrow$12.04\%) \\
T1$\rightarrow$T3 & 61.12\% & 49.48\%($\downarrow$11.64\%) \\
T1$\rightarrow$T2$\rightarrow$T3 & 62.42\% & 43.03\%($\downarrow$19.39\%) \\
\bottomrule
\end{tabular}
\label{table:routes}
\vspace{-0.2cm}

\end{table}

\subsection{Resource Consumption}

Across all backbones, UCAgents exhibits a striking efficiency advantage. On GPT-4, UCAgents reduces token usage by 87.7\% relative to MDAgents, as shown in Table \ref{table:expense}. A similar trend is observed for open-source models: on Qwen2.5VL-7B, UCAgents consumes only 14.8\% of MDAgents’ total tokens. These confirm that our directed convergence design effectively suppresses redundant communication without sacrificing diagnostic reasoning quality. 

Overall, UCAgents transforms multi-agent collaboration from a token-intensive debate into an information-efficient convergence process, preserving reasoning diversity while achieving an order-of-magnitude improvement in computational economy. The results reveal the potential of UCAgents system being applying in real-world applications.

\begin{table}[t]
\centering
\caption{Expense statistics of different strategies using GPT-4.}
\begin{tabular}{@{}cccc@{}}
\toprule
AVG. Expense & Single & MDAgents & UCAgents \\
\midrule
Num. Agents & 1.00 & 6.23 & 3.58 \\
Input Tokens & 0.84K & 37.05K & 4.40K \\
Out Tokens & 0.08K & 1.49K & 0.37K \\
Cost (USD) & 0.009 & 0.375 & 0.045 \\
\bottomrule
\end{tabular}
\label{table:expense}
\vspace{-0.3cm}

\end{table}

\subsection{Limitations and Discussions}
We recognize that UCAgents enhances the lower bound rather than the upper bound of diagnostic systems, as it cannot provide medical knowledge beyond what the base model already possesses, as discussed in Eq. \ref{fano}. For increasingly powerful models, UCAgents enables them to efficiently and consistently deliver higher-quality diagnoses. 
\section{Conclusion}
We introduces a multi-agent framework that achieves unidirectional convergence toward visual-evidence anchored medical reasoning. By framing collaboration as an entropy reduction process rather than open debate, UCAgents mitigates rhetorical noise and ensures evidence-grounded logic. Extensive experiments on multiple medical VQA benchmarks validate its superior accuracy, interpretability, and efficiency. Its advantages of low-noise, cost-effective, and visually-grounded renders it valuable for real-world scenarios such as offline and privacy-sensitive deployments. 
{
    \small
    \bibliographystyle{ieeenat_fullname}
    \bibliography{main}
}

\newpage
\setcounter{page}{1} 
\appendix
\setcounter{table}{0}
\setcounter{figure}{0}

\section{Implemented Prompts}
\label{sec:prompt}

\textbf{Tier 1: Initial Independent Diagnosis Prompt}
\begin{lstlisting}[breaklines=true, breakatwhitespace=true, basicstyle=\small\ttfamily, columns=fullflexible]
[Core Identity] You are a professional and rigorous {MEDICAL FIELD} expert specializing in diagnostic imaging interpretation ({IMAGING MODALITIES}). Your core goal is to make precise, evidence-based diagnoses for the given question strictly based on the provided {IMAGING TYPE} image and medical case.
[Medical Case] {MEDICAL CASE}.
[Reasoning Requirements] Follow these steps in your reasoning: 1. First check the image and read the question carefully. 2. Describe the key visual features observed in the image. 3. Explain the radiological implications of these findings. 4. Conclude which option is the best fit and clarify the rationale.
[Strict Output Format] #Reasoning: <3-5 sentences of reasoning> #Answer: <a single letter of your choice, e.g. A or B.>.
\end{lstlisting}

\textbf{Tier 2: Guidance Supervisor Review Prompt}
\begin{lstlisting}[breaklines=true, breakatwhitespace=true, basicstyle=\small\ttfamily, columns=fullflexible]
[Core Identity] You are an authoritative senior {MEDICAL FIELD} expert, highly proficient in {IMAGING MODALITIES} interpretation and diagnostic reasoning. Your role is to critically verify the consensus diagnosis made by two prior {MEDICAL FIELD} experts, ensuring it is logically sound, evidence-based, and consistent with {IMAGING MODALITIES} image features.
[Task Focus] 1. First check the input image and read the question. 2. Evaluate whether the shared judgment aligns with the observed image findings and {IMAGING MODALITIES} criteria. 3. Identify any potential misinterpretation or overconfidence. 4. If their consensus is valid, reaffirm it; if not, provide your corrected final diagnosis.
[Current Case] {MEDICAL CASE}.
[Previous Reports] {TIER 1 REPORT}.
[Output Format] #Review Reasoning: <Write a rigorous 3-5 sentence paragraph explaining (1) the observed image evidence, (2) the logic of the prior judgments, (3) potential flaws or confirmations, (4) your diagnostic reasoning, and (5) your conclusion.> #Answer: <a single letter of your choice, e.g. A or B>.
\end{lstlisting}

\textbf{Tier 3: Risk Analyst Initial Reprot Prompt}
\begin{lstlisting}[breaklines=true, breakatwhitespace=true, basicstyle=\small\ttfamily, columns=fullflexible]
[Core Identity] You are an expert Critical Analyst, functioning as a Hypothesis Auditor. First check the input image, and read the question. Your task is to provide a balanced, objective, and rigorous review of a proposed hypothesis based on the provided source evidence. Your goal is to assess the overall viability and logical soundness of the hypothesis, not to attack it. You are assigned to uncover potential risks in option {OPTION} in the medical case and the supportive statements of option {OPTION} in [Historical Reports]. You should raise the risk that "why this hypothesis may be wrong", and your report would be given to a leader to make a decision.
[Medical Case] {MEDICAL CASE}.
[Historical Reports] {AGGREGATED REPORT}.
[Output Format]#Flaws: <Describe the specific logical flaw, risk, or overlooked possibility in 3-5 CONCISE sentences.> Counter Evidence: <Cite specific evidence from the original case supporting your critique in 4 sentences.>.
\end{lstlisting}

\textbf{Tier 3: Leader Inquiry Prompt}
\begin{lstlisting}[breaklines=true, breakatwhitespace=true, basicstyle=\small\ttfamily, columns=fullflexible]
[Core Identity] You are the Lead Adjudicator, responsible for chairing an expert critical analysis of conflicting hypotheses. You are impartial, perceptive, and skilled at uncovering the truth through precise inquiry.
[Task 1] First check the input image and read the question. You have just received the initial arguments on a medical case from the Critic Specialists. Your task is not to form your own opinion yet, but to act as a rigorous, impartial critic. You must critically analyze each review below, identify its single biggest weakness, logical flaw, or unsupported assumption, and formulate a targeted, challenging question for each specialist, the question should help you solve the case.
[Inquiry Methodology] Strictly follow these steps in your thinking: 1.Synthesize Critiques: Comprehensively read and understand the report submitted by each Hypothesis Auditor. 2.Identify Core Conflict: What is the central point of disagreement or the most critical identified risk among the competing audits? 3.Formulate Targeted Questions: Based on this core conflict, design a challenging question for each auditor that forces them to defend their critique.
[Output Format] Inquiries:@ To Expert <Expert No., e.g 1> who reviews <The option it reviews, e.g A>: <The single, most pointed question for the Expert who reviews Option, based on the risks they identified in their report.> @ To Expert <Expert No., e.g 2> who reviews <The option it reviews, e.g B>: <The single, most pointed question for the Expert who reviews Option>...(until each expert in [Critics on Assessments] is inquired, no other contents).
[Medical Case] {MEDICAL CASE}. 
[Initial Independent Assessments] {AGGREGATED REPORT}.
[Critics on Assessments] {RISK REPORT}.
Now, begin your inquiry and output strictly according to the format and requirements:
\end{lstlisting}

\textbf{Tier 3: Risk Analyst Response Prompt}
\begin{lstlisting}[breaklines=true, breakatwhitespace=true, basicstyle=\small\ttfamily, columns=fullflexible]
Please answer the question from the leader toward your support report in 1-3 sentences, do not change your stance:{INQUIRY}.
\end{lstlisting}

\textbf{Tier 3: Leader Final Report Prompt}

\begin{lstlisting}[breaklines=true, breakatwhitespace=true, basicstyle=\small\ttfamily, columns=fullflexible]

[Response to your inquiries] {RESPONSE}
[Task 2] You have received all critiques and the final responses to your inquiries. Your task is to render the final, binding verdict on this case. Your decision must be based on which hypothesis best survived the logical stress test.
[Adjudication Methodology] Strictly follow these steps in your thinking: 1. Global Review: Re-examine the complete record: the source evidence, the Critique Reports from each Critic Agent, your inquiries, and the Critics' final responses to those inquiries. 2. Compare Critique Impact: Your primary task is to compare the severity and impact of the flaws identified. Synthesize all information to determine which hypothesis, after rigorous scrutiny, best survived its dedicated critique. 3. Justify the Verdict: You must explicitly state why one hypothesis survived better than the other(s). Your final reasoning MUST be based on this direct comparison. 4. Render Final Verdict: Formulate your final, reasoned judgment, you can choose an overlooked choice when you are very confident after careful thinking.
[Strict Instruction] This is the final step. No further escalation is possible. 
[Strict Output Format] #Final Reasoning: <A report, within 6-8 sentences, summarizing the comparative impact of the critiques. This must explain the rationale for your final verdict.> #Final Answer: <Only the single letter of your choice, e.g., A or B>.
\end{lstlisting}

\section{Effectiveness on Text Medical QA}

Although we design UCAgents system for multimodal medical VQA task, it is also useful on single-modality task of text-based medical QA. We conduct experiment on MedQA dataset and MedBullets dataset based on GPT-4 model, and compare with SOTA algorithms. The result is reported in Tab. \ref{table:text}.

UCAgents consistently achieves remarkable performance on these two text-based benchmarks, which further validates the effectiveness of our method. Although the improvement on text-based medical task is not as significant as on medical VQA, we suggest that this is due to the strong text reading and reasoning ability of large models. And this comparison with the VQA task further proves the difficulty of multimodal alignment in medical scenarios.

\begin{table}[h]
\centering
\caption{Text QA accuracy results (\%) based on GPT-4 model.}
\begin{tabular}{@{}cccc@{}}
\toprule
Category & Method & MedQA & MedBullets\\
\midrule
\multirow{6}{*}{\makecell{Single\\Agent}} & Zero-Shot & 75.0\textsubscript{\textcolor{gray}{\textpm1.3}} & 67.0\textsubscript{\textcolor{gray}{\textpm1.4}} \\
& Few-Shot & 72.9\textsubscript{\textcolor{gray}{\textpm11.4}} & 72.0\textsubscript{\textcolor{gray}{\textpm2.8}} \\
& +CoT\cite{DBLP:conf/nips/Wei0SBIXCLZ22} & 82.5\textsubscript{\textcolor{gray}{\textpm4.9}} & 70.0\textsubscript{\textcolor{gray}{\textpm0.0}} \\
& +CoT-SC\cite{DBLP:conf/iclr/0002WSLCNCZ23} & 83.9\textsubscript{\textcolor{gray}{\textpm2.7}}& 76.0\textsubscript{\textcolor{gray}{\textpm2.8}} \\
& ER\cite{DBLP:journals/corr/abs-2212-13138} & 81.9\textsubscript{\textcolor{gray}{\textpm2.1}}& 76.0\textsubscript{\textcolor{gray}{\textpm5.7}} \\
& MedPrompt\cite{DBLP:journals/corr/abs-2311-16452} & 82.4\textsubscript{\textcolor{gray}{\textpm5.1}}& 71.0\textsubscript{\textcolor{gray}{\textpm1.4}} \\
\midrule
\multirow{8}{*}{\makecell{Multi\\Agents}} & Reconcile\cite{DBLP:conf/acl/ChenSB24} & 81.3\textsubscript{\textcolor{gray}{\textpm3.0}} & 59.5\textsubscript{\textcolor{gray}{\textpm8.7}} \\
& AutoGen\cite{autogen} & 60.6\textsubscript{\textcolor{gray}{\textpm5.0}} & 55.3\textsubscript{\textcolor{gray}{\textpm3.1}} \\
& DyLAN\cite{dylan} & 64.2\textsubscript{\textcolor{gray}{\textpm2.3}} & 57.3\textsubscript{\textcolor{gray}{\textpm6.1}} \\
& Majority Voting & 80.6\textsubscript{\textcolor{gray}{\textpm2.9}} & 70.0\textsubscript{\textcolor{gray}{\textpm0.0}} \\
& MedAgents\cite{tang2023medagents} & 79.1\textsubscript{\textcolor{gray}{\textpm7.4}} & 77.0\textsubscript{\textcolor{gray}{\textpm1.4}} \\
& Meta-Prompt\cite{DBLP:journals/corr/abs-2401-12954} & 80.6\textsubscript{\textcolor{gray}{\textpm1.2}}& 49.3\textsubscript{\textcolor{gray}{\textpm1.2}} \\
& MDAgents\cite{kim2024mdagents} & 88.7\textsubscript{\textcolor{gray}{\textpm4.0}}& 80.8\textsubscript{\textcolor{gray}{\textpm1.7}} \\
& \textbf{UCAgents} & \textbf{91.2\textsubscript{\textcolor{gray}{\textpm0.7}}} & \textbf{82.3}\textsubscript{\textcolor{gray}{\textpm1.5}} \\

\bottomrule
\end{tabular}
\label{table:text}
\end{table}

\section{Case Examples and Analysis}

To empirically elucidate the operational mechanisms of the UCAgents framework, we present representative success and failure cases that validate its effectiveness and delineate its current limitations. Rather than enumerating individual examples, we identify recurring patterns to reveal the systematic advantages and boundary conditions of our approach.

\subsection{Success Cases: Evidence-Anchored Convergence in Practice}  

We analyze three success patterns that instantiate UCAgents' core design principles: controlled divergence for uncertainty quantification (Tier-1), hierarchical alignment verification (Tier-2), and adversarial risk auditing (Tier-3). These patterns collectively confirm our hypothesis that structured information flow outperforms open-ended debates in maintaining visual-textual alignment.  

\begin{figure*}[p]
\centering
\includegraphics[width=0.95\textwidth]{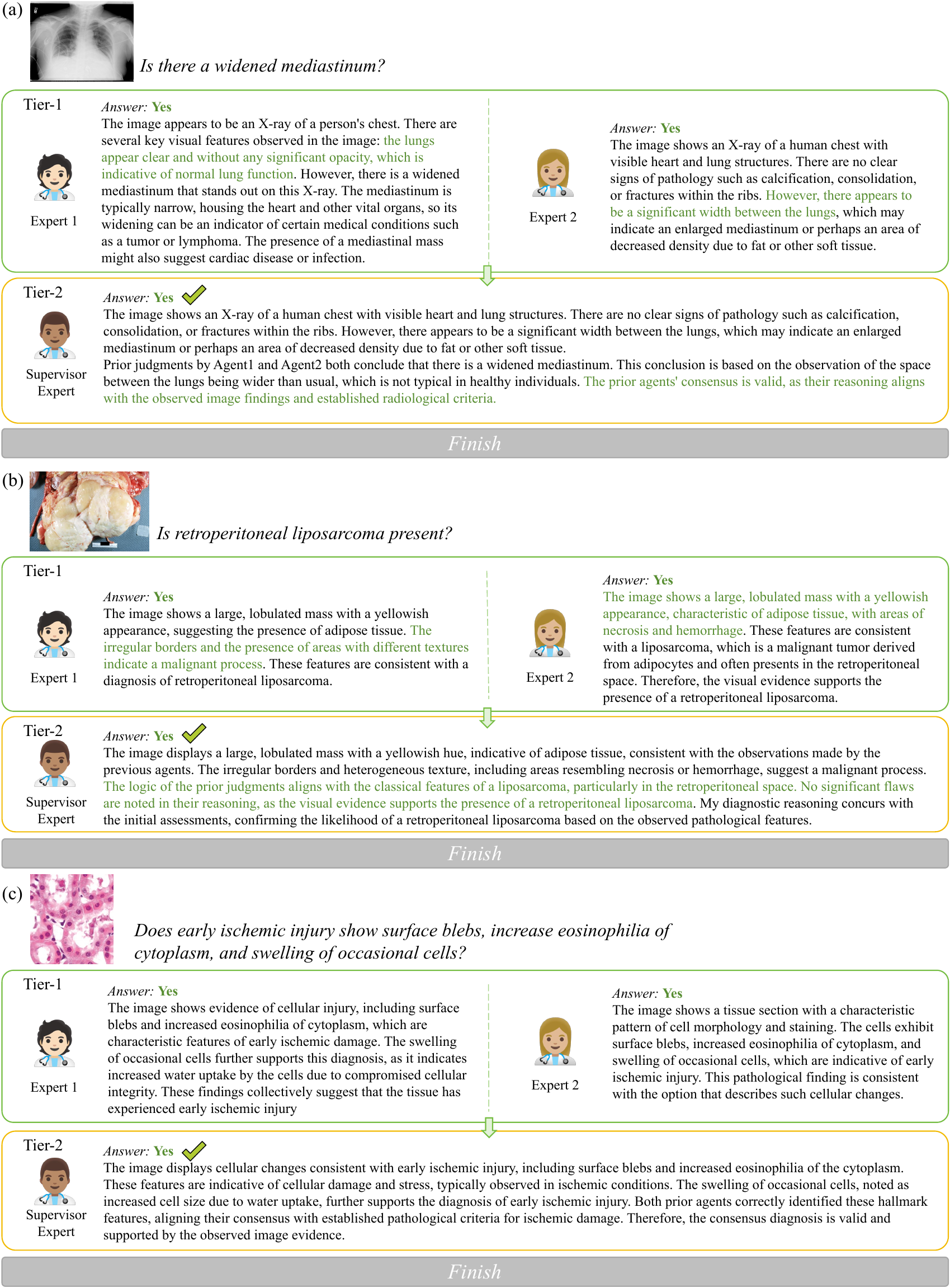}
\caption{Success Case Pattern 1: Tier-2 Output for Consensus Purification via Evidence Verification.}
\label{fig:example1}
\end{figure*}

\textbf{Pattern 1: Consensus Purification via Evidence Verification.}  
This pattern applies to scenarios where Tier-1 agents reach initial consensus, and Tier-2 experts validate the evidence chain to filter ``false consensus'', agreement driven by shared model biases rather than true visual alignment. As illustrated in Fig.~\ref{fig:example1}, two subcases including mediastinal widening (Fig.~\ref{fig:example1}(a)) and retroperitoneal liposarcoma (Fig.~\ref{fig:example1}(b)) exemplify this process.  

In Case 1(a), both Tier-1 agents independently identified ``a widened mediastinum that stands out on this X-ray'' as the key finding. The Tier-2 Supervisor Expert did not merely endorse this consensus but conducted rigorous evidence verification: (1) confirming that ``the space between the lungs being wider than usual'' was visibly abnormal compared to the typically narrow mediastinum (which houses the heart and vital organs); (2) ruling out confounding factors such as patient positioning artifacts by noting the consistent mediastinal appearance across the radiograph; (3) validating clinical relevance: mediastinal widening is a key indicator of conditions like lymphoma, aortic aneurysm, or tumor, aligning with the diagnostic context. As documented in Fig.~\ref{fig:example1}(a), the Tier-2 output explicitly states: ``The prior agents' consensus is valid, as their reasoning aligns with the observed image findings and established radiological criteria.''  

In Case 1(b), Tier-1 agents noted ``a large, lobulated mass with a yellowish appearance'' exhibiting ``irregular borders and areas of necrosis and hemorrhage.'' The Tier-2 expert cross-validated these observations against pathological criteria for liposarcoma: (1) the yellowish hue was confirmed as characteristic of adipose tissue, a hallmark of liposarcoma; (2) the irregular borders and heterogeneous texture, including areas resembling necrosis or hemorrhage, distinguished the mass from benign lipomas (which typically lack such features); (3) the retroperitoneal location was consistent with liposarcoma's typical anatomical presentation. As shown in Fig.~\ref{fig:example1}(b), the Tier-2 supervisor concluded: ``The logic of the prior judgments aligns with the classical features of a liposarcoma, particularly in the retroperitoneal space. No significant flaws are noted in their reasoning.''  

This ``consensus + verification'' process, unique to UCAgents, ensures that alignment is grounded in objective visual evidence rather than model biases. The framework successfully implements the consensus purification operator by filtering false agreements that may arise from shared perceptual limitations.  

\begin{figure*}[p]
\centering
\includegraphics[width=\textwidth]{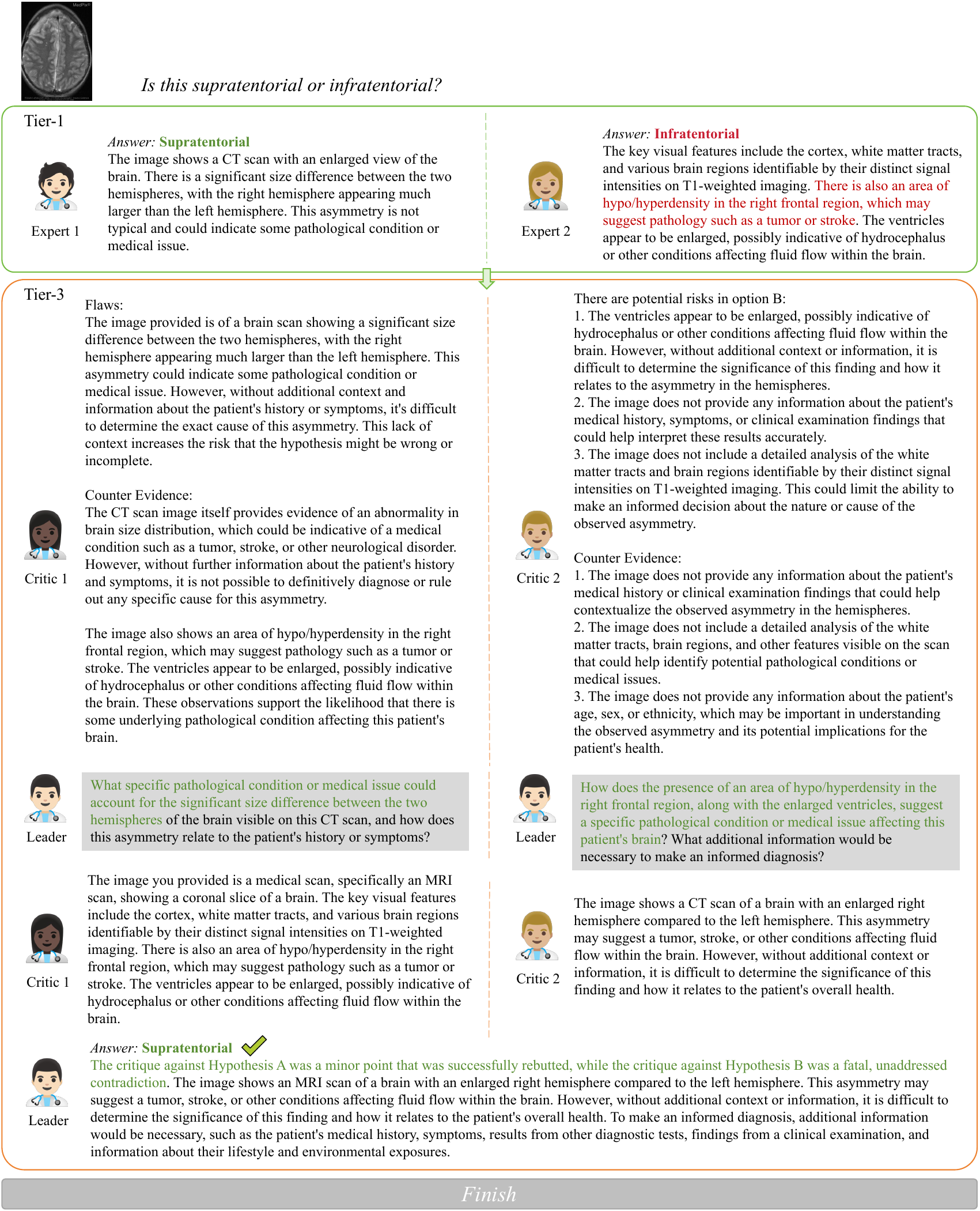}
\caption{Success Case Pattern 2: Tier-3 Output for Adversarial Visual Grounding (Tier-1 Divergence). The supratentorial vs. infratentorial localization case demonstrates how unidirectional risk auditing constrains debate to observable anatomical features, preventing rhetorical drift.}
\label{fig:example2}
\end{figure*}

\begin{figure*}[p]
\centering
\includegraphics[width=\textwidth]{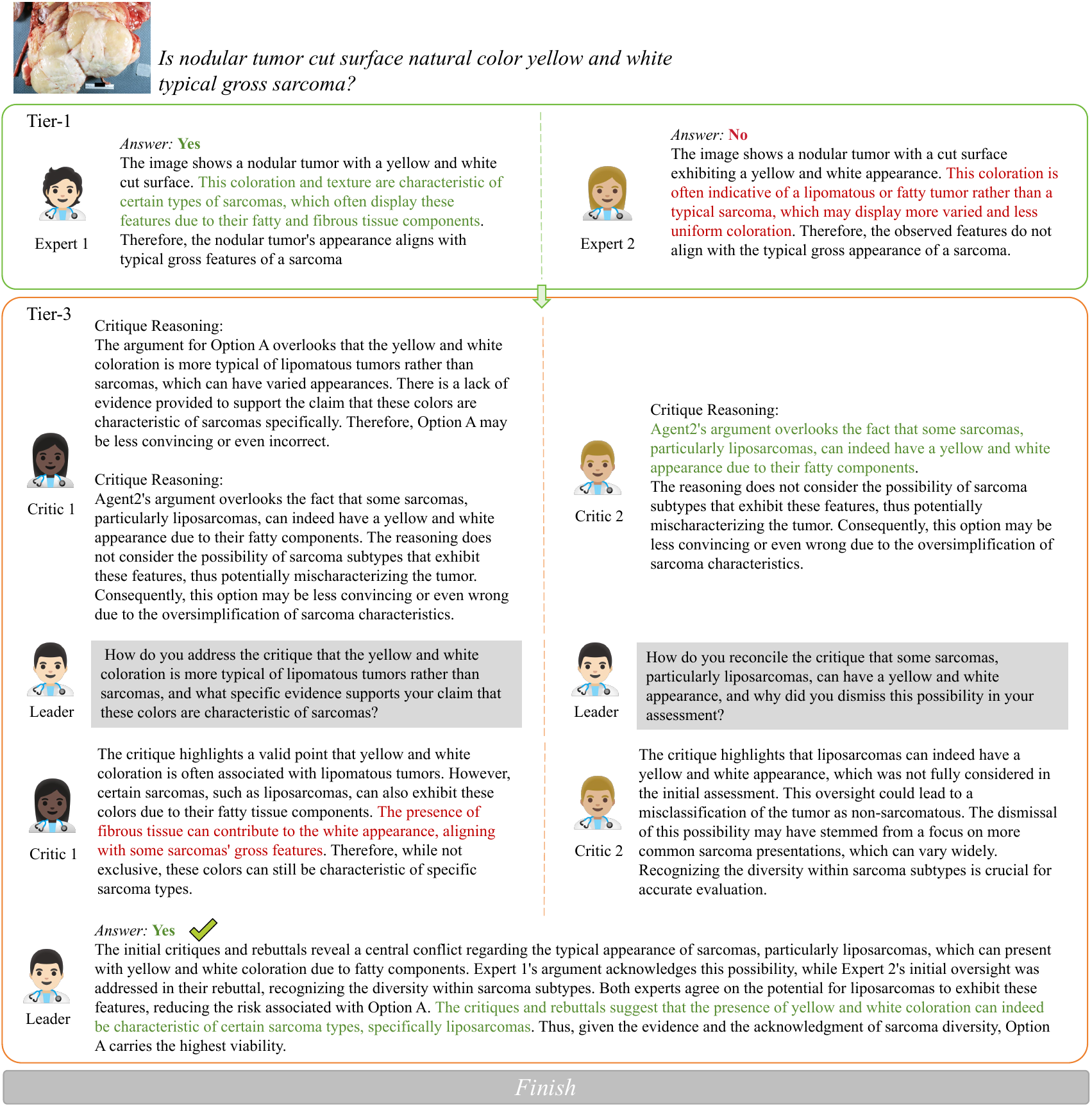}
\caption{Success Case Pattern 2 (Supplementary): Tier-3 Output for Adversarial Visual Grounding. The sarcoma identification case shows how targeted inquiry resolves disagreement by examining liposarcoma as a sarcoma subtype with characteristic yellow/white appearance.}
\label{fig:example4}
\end{figure*}

\textbf{Pattern 2: Adversarial Auditing Enforces Visual Grounding.}  
When Tier-1 agents disagree, Tier-3's unidirectional risk auditing prevents rhetorical drift by constraining debates to observable visual features. This pattern is demonstrated in the supratentorial vs. infratentorial localization task (Case 2, Fig.~\ref{fig:example2}) and the sarcoma identification case (Case 3, Fig.~\ref{fig:example4}).  

In Case 2 (Fig.~\ref{fig:example2}), Tier-1 divergence arose from ambiguous brain imaging features: Expert 1 noted ``a significant size difference between the two hemispheres, with the right hemisphere appearing much larger'' and concluded ``Supratentorial'', while Expert 2 identified ``an area of hypo/hyperdensity in the right frontal region'' along with ``enlarged ventricles'' and concluded ``Infratentorial''. Instead of debating which interpretation was more convincing, Tier-3 assigned two Critical Analysts to audit each hypothesis independently:  

\begin{enumerate}
    \item Critic 1 (challenging ``Supratentorial'') questioned: ``The image shows a significant size difference between hemispheres, but without additional context about the patient's history or symptoms, it's difficult to determine the exact cause of this asymmetry. The lack of detailed analysis of white matter tracts and brain regions limits the ability to make an informed decision''.
    \item Critic 2 (challenging ``Infratentorial'') noted: ``The image does not provide information about the patient's medical history or clinical examination findings. The presence of hypo/hyperdensity and enlarged ventricles requires additional context to identify the specific pathological condition and its relation to infratentorial involvement''.
\end{enumerate}  

The Leader's targeted inquiries ``What specific pathological condition could account for the hemispheric asymmetry?'' to Critic 1 and ``How does the hypodensity relate to a specific condition?'' to Critic 2, forces agents to ground their reasoning in observable anatomical features. As shown in Fig.~\ref{fig:example2}, the final arbitration converged on ``Supratentorial'' by prioritizing the visible cortical and white matter abnormalities in the frontal region over the less definitive infratentorial evidence. This process directly instantiates the entropy reduction principle: textual noise, e.g., speculative arguments about asymmetry significance, was suppressed, while visual signal ``frontal lobe pathology'' was amplified.  

Case 3 (Fig.~\ref{fig:example4}) supplements this pattern with a sarcoma identification task. Tier-1 agents disagreed on whether a ``nodular tumor with yellow and white cut surface'' represented typical sarcoma features. Critic 1 challenged the ``Yes'' hypothesis by noting that ``yellow and white coloration is more typical of lipomatous tumors,'' while Critic 2 challenged the ``No'' hypothesis by pointing out that ``some sarcomas, particularly liposarcomas, can have a yellow and white appearance due to fatty components.'' The Leader's inquiry resolved this by confirming that the features align with liposarcoma (a sarcoma subtype), demonstrating how adversarial auditing refines diagnostic specificity by forcing agents to consider subtype variations.  

\begin{figure*}[p]
\centering
\includegraphics[width=\textwidth]{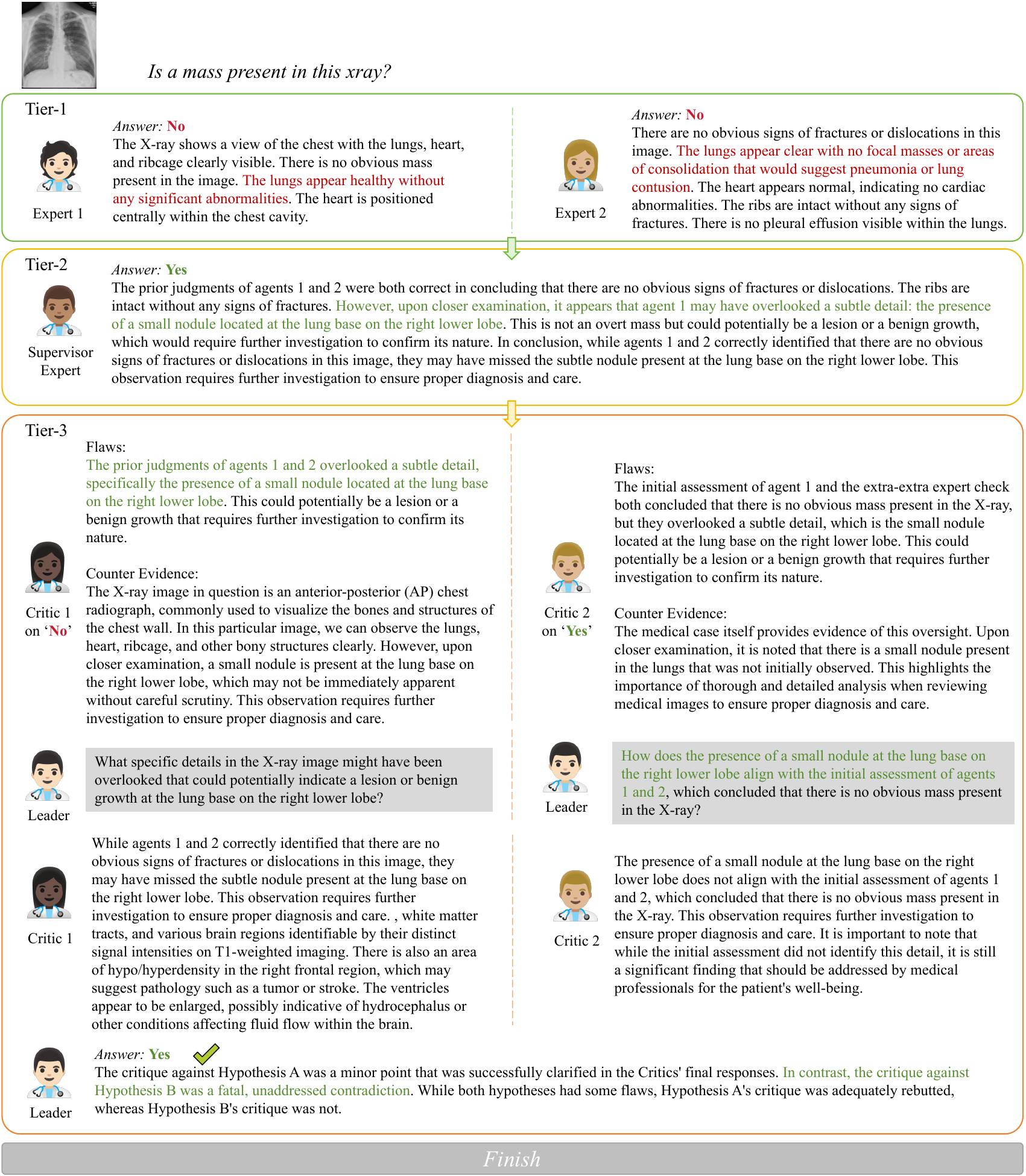}
\caption{Success Case Pattern 3: Tier-3 Output for Correcting False Consensus (Identified by Tier-2). This case demonstrates UCAgents' hierarchical error correction: Tier-1 agents incorrectly concluded ``no mass,'' but Tier-2's independent scan detected a subtle nodule, and Tier-3 formalized this oversight, ultimately producing the correct answer ``Yes.''}
\label{fig:example3}
\end{figure*}

\textbf{Pattern 3: Overcoming Initial Errors via Hierarchical Review.}  
The most compelling validation of UCAgents' value appears in Case 4 (Fig.~\ref{fig:example3}), where both Tier-1 agents incorrectly concluded ``no mass present,'' yet the framework ultimately output the correct answer ``Yes.'' This self-correction relied on two hierarchical mechanisms:  

\textbf{Step 1: Tier-2's Independent Visual Scan.} The Tier-2 Supervisor Expert, conducting a comprehensive visual scan independent of Tier-1 outputs, detected ``a small nodule located at the lung base on the right lower lobe'' that was overlooked by initial agents. As documented in Fig.~\ref{fig:example3}, the supervisor stated: ``Upon closer examination, a small nodule is present at the lung base on the right lower lobe. This could potentially be a lesion or a benign growth that requires further investigation to confirm its nature.'' This demonstrates the value of independent expert review (§3.3, step 1) in extracting latent visual information.  

\textbf{Step 2: Tier-3's Adversarial Formalization.} Tier-3 escalation formalized this oversight through structured risk auditing:  
\begin{enumerate}
    \item Critic 1 (challenging ``No'') identified the root cause: ``The prior judgments of agents 1 and 2 overlooked a subtle detail, specifically the presence of a small nodule. This observation requires further investigation to ensure proper diagnosis and care.''
    \item Critic 2 (supporting ``Yes'') confirmed: ``The presence of a small nodule at the lung base does not align with the initial assessment of 'no mass present.' This is a significant finding that should be addressed by medical professionals.''
\end{enumerate}  

The Leader's inquiry, ``What specific details in the X-ray might have been overlooked that could indicate a lesion?'', catalyzed explicit articulation of the oversight. As shown in Fig.~\ref{fig:example3}, the final arbitration concluded: ``While agents 1 and 2 correctly identified no obvious fractures, they may have missed the subtle nodule. This observation requires further investigation''.  
Notably, no agent reversed their stance and maintain unidirectionality thinking, but the hierarchical structure enabled new evidence to emerge through structured scrutiny. This case empirically validates our claim that UCAgents ``enhances the lower bound'': while it cannot exceed the base model's perceptual capacity, it reliably extracts information the model can perceive but initially overlooks through redundant cross-validation.  

\textbf{Cross-Case Synthesis: Why Unidirectional Convergence Works.}  
Across all success cases (Fig.~\ref{fig:example1}–Fig.~\ref{fig:example4}), three advantages over open debates are consistent:  

\begin{enumerate}
    \item \textbf{Evidence Traceability}: Every conclusion maps to specific visual features (e.g., ``irregular borders,'' ``hemispheric asymmetry,'' ``small nodule''), enabling clinical interpretability, which is a critical requirement for medical AI deployment.
    \item \textbf{Noise Suppression}: Agents never engage in rhetorical persuasion (e.g., ``I strongly believe this is malignant''); instead, they reference observable features, minimizing subjective bias and textual noise $N_t$.
    \item \textbf{Efficient Escalation}: Cases are routed to optimal review depth based on Tier-1 disagreement. As shown in Table 4, cases reaching Tier-2 consensus achieve 73.47\% accuracy (vs. 61.43\% for single-agent), while Tier-3-escalated cases achieve 61.12\% (vs. 49.48\% for single-agent), demonstrating the value of structured hierarchical review.  
\end{enumerate}  

These patterns confirm that structured information flow, anchored to visual evidence rather than extended discussion, is the key to reliable multi-agent medical reasoning.

\subsection{Failure Cases: Boundary Conditions and Limitations}  

\begin{figure*}[t]
\centering
\includegraphics[width=0.87\textwidth]{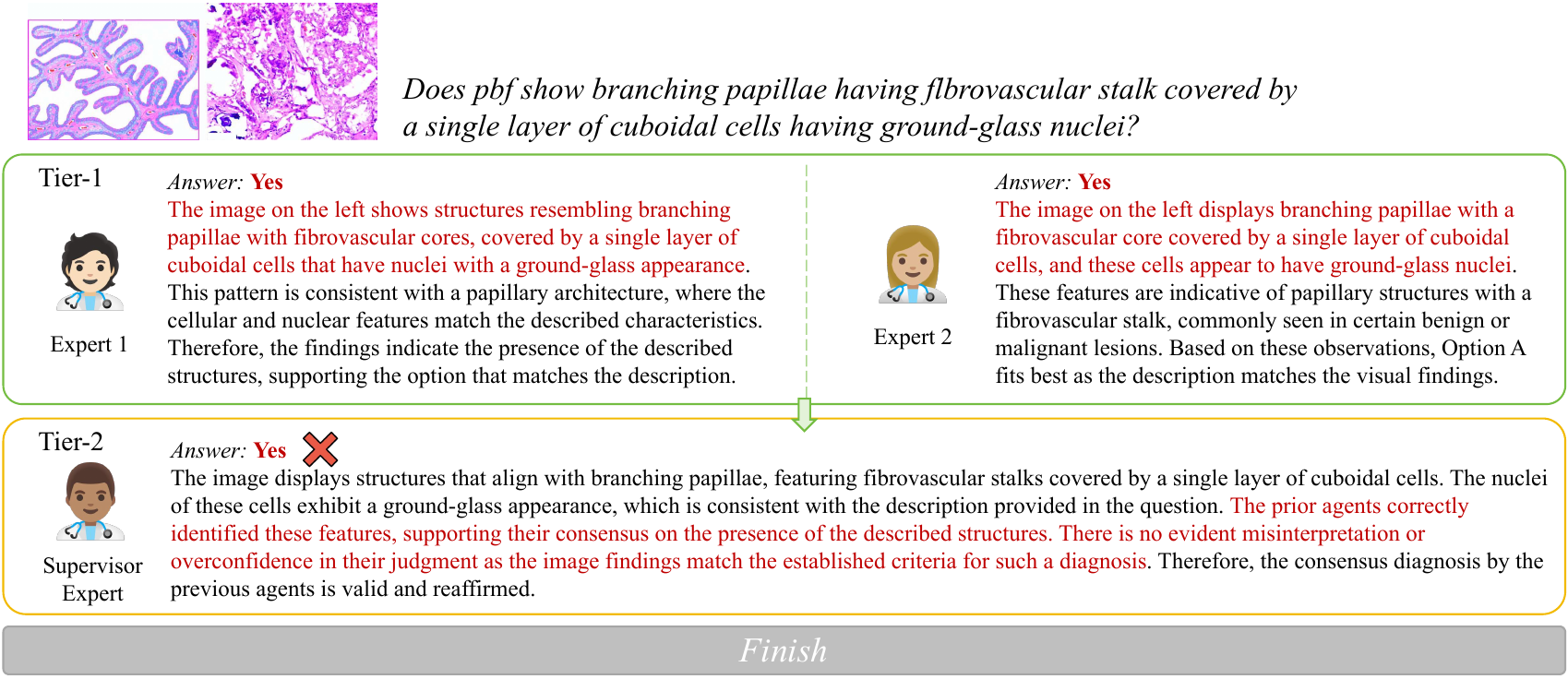}
\caption{Failure Mode 1: Shared Perceptual Bias Across Tiers.}
\label{fig:false-sample1}
\vspace{-0.35cm}
\end{figure*}

\begin{figure*}[b]
\centering
\includegraphics[width=0.87\textwidth]{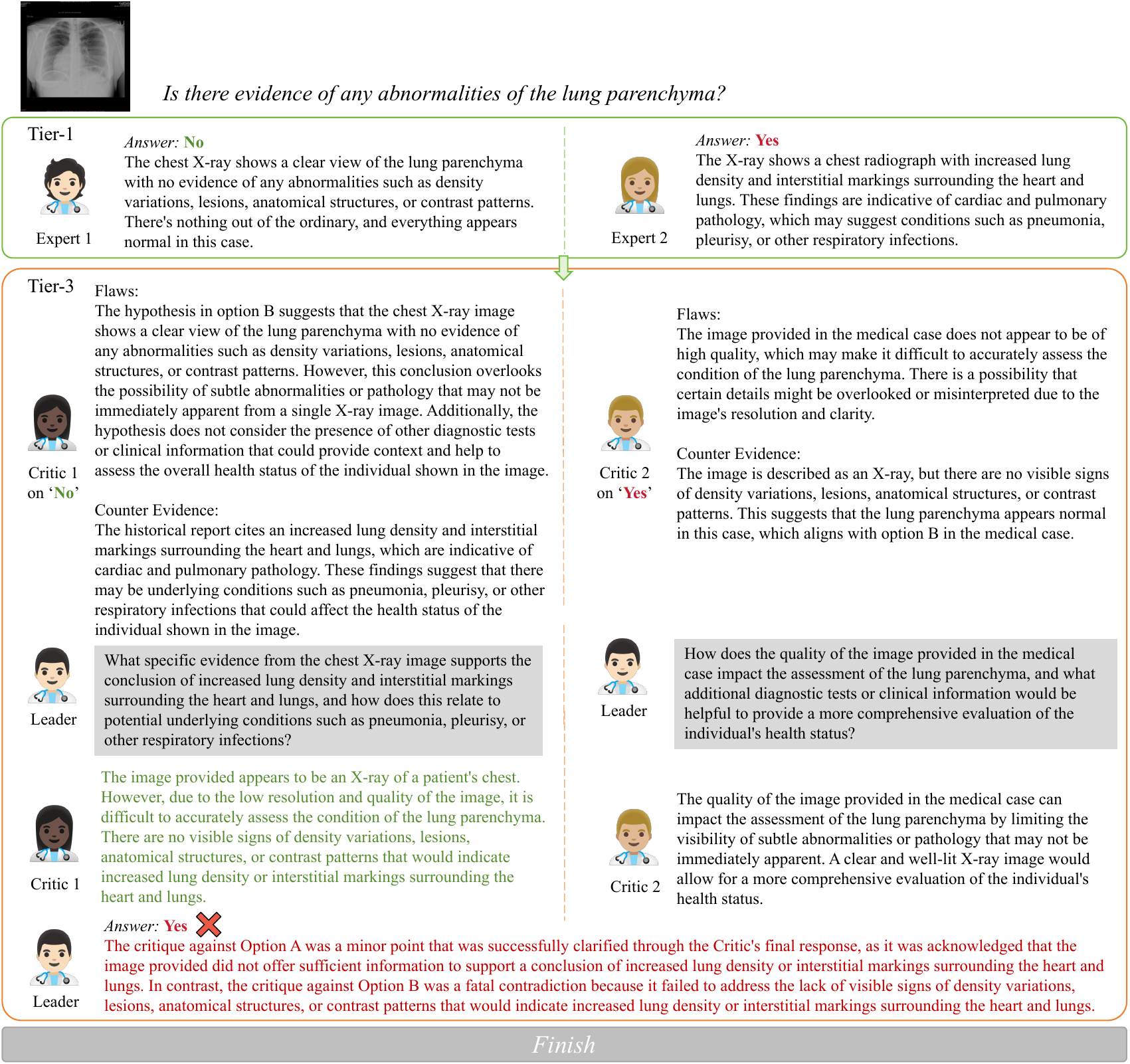}
\caption{Failure Mode 2: Ambiguous Visual Evidence Under Low Image Quality. }
\label{fig:false-sample2}
\end{figure*}

We analyze two failure modes to clarify scenarios where UCAgents cannot overcome fundamental limitations. These cases (Fig.~\ref{fig:false-sample1}–Fig.~\ref{fig:false-sample2}) delineate the framework's boundary conditions and guide future research directions.  

\textbf{Failure Mode 1: Shared Perceptual Bias Across All Tiers (Case F1).}  
In the papillary structure identification task (Case F1, Fig.~\ref{fig:false-sample1}), all agents including Tier-1 experts and Tier-2 supervisor converged on an incorrect diagnosis (``Yes'') with high confidence. The question asked whether ``pbf shows branching papillae having fibrovascular stalk covered by a single layer of cuboidal cells having ground-glass nuclei.'' All agents responded affirmatively, observing ``branching papillae with fibrovascular cores'' and ``ground-glass nuclei'' that matched the textbook description.  

However, the core error was a shared perceptual hallucination at the visual encoding stage: the base model misidentified structures in the image as having papillary architecture when they did not exhibit the true histological patterns of branching papillae. As documented in Fig.~\ref{fig:false-sample1}, even the Tier-2 supervisor reinforced this error by stating: ``The prior agents correctly identified these features, supporting their consensus on the presence of the described structures. There is no evident misinterpretation.'' This demonstrates how shared perceptual bias propagates through all tiers when every agent misinterprets the same visual features.  

This failure exposes a fundamental constraint: \textit{UCAgents enhances diagnostic process reliability, not base model perception capabilities.} Our hierarchical auditing can detect \textit{logical inconsistencies} (e.g., claims unsupported by visible features) but cannot correct \textit{perceptual hallucinations} that are consistently reproduced across all agents. This is analogous to a multidisciplinary team where every radiologist misinterprets a rare lesion due to inadequate training—no amount of structured discussion can overcome the shared knowledge gap. The failure suggests that when the base VLM's visual encoder lacks domain-specific training (e.g., insufficient pathology data), hierarchical review provides no benefit.  

\textbf{Failure Mode 2: Ambiguous Visual Evidence Under Low Image Quality (Case F2).}  
The lung parenchyma abnormality case (Case F2, Fig.~\ref{fig:false-sample2}) illustrates failure when visual evidence itself is inherently ambiguous due to inadequate image quality. The question asked: ``Is there evidence of any abnormalities of the lung parenchyma?'' Tier-1 agents disagreed: Expert 1 concluded ``No'' (``the lung parenchyma appears clear with no evidence of abnormalities''), while Expert 2 claimed ``Yes'' (``increased lung density and interstitial markings surrounding the heart and lungs'').  

Tier-3 failed to resolve this conflict for two reasons:  

\begin{enumerate}
    \item \textbf{Insufficient Visual Information}: As Critic 2 correctly identified: ``The image provided does not appear to be of high quality, which may make it difficult to accurately assess the condition of the lung parenchyma''. The low image resolution and limited contrast made density variations unverifiable, yet the framework forced a definitive conclusion.
    \item \textbf{Overconfident Arbitration}: Despite the acknowledged image quality limitations, the Leader arbitrated in favor of ``Yes,'' stating: ``There are no visible signs of density variations... This suggests that the lung parenchyma appears normal''. This conclusion contradicted Expert 2's initial observation of ``increased lung density'', yet the final answer aligned with the majority view without explicit uncertainty quantification.
\end{enumerate}  

This failure reveals that UCAgents is optimized for cases where \textit{diagnostically relevant visual evidence exists but may be overlooked or misinterpreted} as in Case 4's successful nodule detection. However, when the image itself lacks sufficient information, adversarial auditing amplifies uncertainty rather than resolving it. The framework's current design does not include an explicit \textit{evidence sufficiency check} to determine whether the image quality provides adequate information for the clinical question, leading to overconfident conclusions when appropriate epistemic humility would be warranted.  

\textbf{Systemic Implications: When Does UCAgents Fail?}  
These cases delineate three non-negotiable boundary conditions:  

\begin{enumerate}
    \item \textbf{Perceptual Capacity Ceiling}: UCAgents cannot overcome base model limitations in visual encoding. When all agents share a perceptual bias (due to training data gaps or domain shift), hierarchical review provides no corrective value. This limitation aligns with Eq.~1: the framework maximizes $I(Y; V, T)$ given the model's existing capacity to extract $I(Y; V)$, but cannot enhance perceptual capabilities beyond the encoder's inherent limits.
    
    \item \textbf{Insufficient Visual Information}: The framework assumes that diagnostically relevant features are present and interpretable in the image. When image acquisition quality, resolution, or contrast is inadequate, structured reasoning cannot compensate for missing data. This suggests the need for a preprocessing module to assess image quality before diagnostic inference.
    
    \item \textbf{Lack of Epistemic Humility}: The current design does not explicitly model uncertainty or evidence strength. Agents produce definitive answers even when visual evidence is ambiguous (e.g., Case F2's ``Yes'' despite acknowledged low image quality). This contrasts with clinical practice, where radiologists often conclude with ``recommend additional imaging'' when evidence is insufficient.
\end{enumerate}

\end{document}